\documentclass[12pt]{article}

\usepackage{sbc-template}
\usepackage{graphicx,url}
\usepackage[utf8]{inputenc}
\usepackage[brazil]{babel}

\usepackage{booktabs}
\usepackage{multirow}
\usepackage{amsmath}
\usepackage{booktabs}
\usepackage{color, colortbl}

\title{How effective are Graph Neural Networks in Fraud Detection for Network Data?}

\author{Ronald D. R. Pereira\inst{1}, Fabrício Murai\inst{1}}

\address{Departamento de Ciência da Computação\\
  Universidade Federal de Minas Gerais (UFMG) -- Belo Horizonte, MG -- Brasil
  \email{ronald.pereira@dcc.ufmg.br, murai@dcc.ufmg.br}}

\begin{document} 

\maketitle

% Temos que reduzir esses abstract em até 10 linhas no máximo

\begin{abstract}
Graph-based Neural Networks (GNNs) are recent models created for learning representations of nodes (and graphs), which have achieved promising results when detecting patterns that occur in large-scale data relating different entities. Among these patterns, financial fraud stands out for its socioeconomic relevance and for presenting particular challenges, such as the extreme imbalance between the positive (fraud) and negative (legitimate transactions) classes, and the concept drift (i.e., statistical properties of the data change over time). Since GNNs are based on message propagation, the representation of a node is strongly impacted by its neighbors and by the network's hubs, amplifying the imbalance effects. Recent works attempt to adapt undersampling and oversampling strategies for GNNs in order to mitigate this effect without, however, accounting for concept drift. In this work, we conduct experiments to evaluate existing techniques for detecting network fraud, considering the two previous challenges. For this, we use real data sets, complemented by synthetic data created from a new methodology introduced here. Based on this analysis, we propose a series of improvement points that should be investigated in future research.
\end{abstract}
     
\begin{resumo} 
Redes Neurais baseadas em Grafos (GNNs) são modelos recentes criados para o aprendizado de representações de nós (e de grafos), que alcançaram resultados promissores na detecção de padrões que ocorrem em dados de larga escala que relacionam diferentes entidades. Dentre esses padrões, fraudes financeiras se destacam por sua relevância socioeconômica e por apresentarem desafios particulares, tais como o desbalanceamento extremo entre as classes positivas (fraudes) e negativas (transações legítimas), e o desvio de conceito (i.e., propriedades estatísticas dos dados mudam ao longo do tempo). Como as GNNs são baseadas em propagação de mensagem, a representação de um nó acaba sendo muito impactada pelos seus vizinhos e pelos hubs da rede, amplificando os efeitos do desbalanceamento. Pesquisas recentes tentam adaptar estratégias de subamostragem e sobreamostragem para GNNs a fim de mitigar esse efeito sem, contudo, considerar o desvio de conceito. Neste trabalho, realizamos uma série de experimentos para avaliar técnicas existentes de detecção de fraudes em rede, considerando os dois desafios anteriores. Para isso, utilizamos conjuntos de dados reais, complementados por dados sintéticos criados a partir de uma nova metodologia introduzida aqui. Baseado nessa análise, propomos uma série de pontos de melhoria a serem investigados em pesquisas futuras.
\end{resumo}

\section{Introdução}

Transações financeiras tornaram-se muito mais comuns nos últimos anos com o aumento de criptomoedas, bancos digitais e \textit{gateways} de pagamento online. Essas tecnologias transferiram o controle dos bancos para os usuários, tornando o processo de compra e transferência mais distribuído e acessível para a população em geral, mas também mais suscetível a fraudes. Esse problema afeta as métricas e os resultados financeiros de muitas empresas, já que o valor total da fraude às vezes pode ultrapassar sua receita ou tornar a margem de lucro tão pequena (mesmo que o valor da fraude seja pequeno quando comparado a transações legítimas) e, a prazo, tornar a falência apenas uma questão de tempo.

Muitas pesquisas já foram feitas nesta área, mas a maioria delas se concentra na detecção de transações fraudulentas com cartão de crédito. O uso de modelos de aprendizagem supervisionada amplamente conhecidos, como Árvores de Decisão e \textit{Support Vector Machine} (SVM) \cite{csahin2011detecting}, Redes Neurais Artificiais e Modelos Gráficos Probabilísticos \cite{maes2002credit}, e até mesmo técnicas de mineração de dados \cite{brause1999neural, chawla2009data}, tem sido um traço comum neste campo de pesquisa.

A detecção de fraudes tem se mostrado um domínio muito mais complexo do que pode-se imaginar à primeira vista \cite{abdallah2016fraud}. O projeto de sistemas de antifraude tem como desafios adicionais, em relação a sistemas de previsão tradicionais, limitar o número de alarmes falsos (ou seja, evitar uma alta taxa de falsos-positivos) e lidar com a diminuição na precisão das previsões conforme seus dados de treinamento se tornam obsoletos, devido ao desvio de conceito. Em mineração de dados e aprendizado de máquina, o desvio de conceito se refere ao fenômeno de ter a distribuição de dados subjacentes mudando ao longo do tempo, o que torna muito difícil manter a qualidade das previsões. Outro desafio é a natureza desequilibrada dos conjuntos de dados neste problema, pois há muito menos fraude do que instâncias legítimas, o que torna ainda mais difícil para as técnicas de aprendizado supervisionado inferir a distribuição intrínseca dos dados.

Recentemente, uma importante relação entre entidades envolvidas em transações financeiras que vai além das relações de primeira ordem ``origem-destino'', foi incorporada a técnicas de aprendizagem de grafos para detectar fortes relações entre instâncias distintas de dados por meio de pesos em arestas usando grafos \cite{weber2018scalable, wagner2019latent, weber2019anti}, mas nenhuma deles tenta mitigar o desbalanceamento do conjunto de dados ou os problemas de desvio de conceito. Esses problemas são abordados em outros trabalhos de pesquisa, mas sem que se tente detectar essas relações usando modelos estruturados de grafos \cite{wang2017imbalanced, wang2020ocgnn, yang2009ensemble, gao2020one, shamshirband2014anomaly}. Passos iniciais foram dados recentemente para solucionar o problema de desvio de conceito usando redes convolucionais de grafos sem, no entanto, considerar o problema de classes desequilibradas em suas avaliações \cite{pareja2020evolvegcn}. Por outro lado, outra pesquisa recente tenta mitigar o problema do desbalanceamento dos dados também usando redes neurais baseadas em grafos, mas sem levar em conta a desvio de conceito \cite{liu2021pickandchoose}.

A dificuldade de desenvolver uma pesquisa com utilização de aprendizagem de grafos em dados de transações financeiras e compará-la com outras reside na escassa disponibilidade de conjuntos de dados públicos que sejam possíveis de serem transformados em uma rede. Nesse contexto, o presente trabalho visa avaliar as técnicas existentes para detecção de fraudes financeiras sob a luz destes dois desafios utilizando-se de conjuntos de dados publicamente disponíveis, sendo esses dois datasets reais - YelpChi \cite{mukherjee2013yelp} e Elliptic Data Set \cite{weber2019anti} - e três gerados sinteticamente por meio de um simulador de transações denominado AMLSim \cite{pareja2020evolvegcn, weber2018scalable}. Mais precisamente, este artigo traz as seguintes contribuições:

\begin{itemize}
    \item Propõe nova metodologia para avaliar a robustez de métodos de detecção de fraudes financeiras baseada em geração de dados sintéticos para uma visão completa dos pontos fortes e fracos;
    \item Quantifica os avanços alcançados por pesquisas recentes na área ao implementá-los e compará-los com modelos anteriormente desenvolvidos;
    \item Encontra e propõe pontos de melhorias para os modelos tidos como estado da arte, de modo a viabilizar futuras pesquisas na área.
\end{itemize}

A partir desses experimentos foi possível concluir que estamos realmente progredindo para melhores desempenhos dada a complexidade inerente do domínio de detecção de fraude em dados em rede. No entanto, a união dos dois maiores desafios da área ainda se mostra pendente, dado que atualmente ainda não foi desenvolvido nenhum modelo ou método que seja capaz de alcançar métricas consistentes lidando com dados altamente desbalanceados e que mudam a sua distribuição gradualmente ao longo do tempo, gerando um fenômeno denominado desvio de conceito.

O restante do artigo está organizado da seguinte forma. Na Seção \ref{formulation} introduzimos a formulação do problema, apresentando também as definições e notações necessárias. Na Seção \ref{related_works} apresentamos os trabalhos relacionados, incluindo os métodos de detecção de fraudes utilizados como base neste estudo. Na Seção \ref{methodology} apresentamos os conjuntos de dados, modelos e métricas empregados nos experimentos. Na Seção \ref{results} apresentamos os resultados gerais dos experimentos realizados. Por último, na Seção \ref{conclusion} expomos as conclusões do trabalho e trabalhos futuros que podem ser utilizados como base em pesquisas posteriores.

\section{Formulação do Problema}
\label{formulation}\label{concept_drift}

Nesta seção, formalizamos o problema de detecção de fraude baseada em grafo, apresentando também as definições e a notação utilizadas.

\noindent
\textbf{Definição 2.1 (Grafo Multi-Relacional).} Dado um grafo $\mathcal{G} = (\mathcal{V}, \mathcal{E}, \mathcal{X}, \mathcal{C})$, onde $\mathcal{V} = \{v_1, \dots, v_N\}$ é o conjunto de vértices (nós); $\mathcal{E} = \{\mathcal{E}_1, \dots, \mathcal{E}_R\}$ é o conjunto de arestas, separadas em $R$ relações; $\mathcal{X}$ e $\mathcal{C}$ são, respectivamente, os conjuntos de características (\textit{features}) e rótulos dos vértices, de modo que para cada nó $v_i \in \mathcal{V}$, $x_i \in \mathcal{X}$ é um vetor de características $d$-dimensional e $c_i \in \mathcal{C}$ é o seu rótulo binário, $\forall \; i=\{1, \dots, N\}$.
\label{def_graph}

\noindent
\textbf{Definição 2.2 (Taxa de Desbalanceamento).} Dado um conjunto de rótulos $\mathcal{C}$ onde $c_i \in \{1,2\}$, podemos definir uma partição $(\mathcal{C}_1,\mathcal{C}_2)$ de  $\mathcal{V}$ tal que $C_k = \{v_i \in \mathcal{V} : v_i = k\}$.
A taxa de desbalanceamento entre $\mathcal{C}_1$ e $\mathcal{C}_2$ é definida como TD $= |\mathcal{C}_1|/|\mathcal{C}_2| \in [0, +\infty)$.
Se TD $> 1$, $\mathcal{C}_1$ é denominada classe majoritária e $\mathcal{C}_2$, classe minoritária. Quando TD $=1$, diz-se que $\mathcal{C}$ é perfeitamente balanceado.

\noindent
\textbf{Definição 2.3 (Desvio de Conceito).} Em análise de dados e aprendizado de máquina, um desvio de conceito ocorre quando as propriedades estatísticas associadas à variável alvo $c$ (rótulo) mudam com o tempo. Isso causa problemas porque as previsões de um dado modelo se tornam menos precisas com o passar do tempo.
O desvio de conceito é bem comum em dados do mundo real, tendo em vista que muitos deles estão diretamente ligados à mapeamentos de comportamentos humanos e da natureza que, por sua vez, não são regidos por nenhuma lei matemática determinística, causando mudanças gradativas em sua distribuição.

\noindent
\textbf{Definição 2.4 (Detecção de Fraude baseada em Grafos).} O problema de detecção de fraude baseada em grafos é definido sobre o grafo multi-relacional $\mathcal{G} = (\mathcal{V}, \mathcal{E}, \mathcal{X}, \mathcal{C})$, da Definição \ref{def_graph}. Cada vértice $v_i$ foi rotulado de forma estática como $c_i$ (fraudulento ou legítimo), mas suas características $x_i$ podem mudar ao longo do tempo.
Desse modo, esse problema consiste em encontrar o rótulo de demais vértices contidos no grafo baseado em seus padrões de transações, com o objetivo de se encontrar similaridades e disparidades entre características de vértices pertencentes à mesma classe e classes distintas, respectivamente.

\section{Trabalhos Relacionados}
\label{related_works}

Os trabalhos relacionados a detecção de fraudes financeiras foram organizados segundo o tipo de abordagem, em aprendizado de máquina tradicional e aprendizado baseado em grafos. Ao final da seção,  discutimos este trabalho no contexto do estado da arte.

\noindent
\textbf{Detecção de fraude por meio de aprendizado de máquina tradicional.} Os primeiros modelos para detecção automática de transações fraudulentas foram propostos há mais de duas décadas \cite{gopinathan1998fraud}. Naquela patente, propõe-se a extração de características discriminativas acerca de transações ilícitas com base em relacionamentos aprendidos entre variáveis conhecidas no domínio de detecção de fraudes. Esses relacionamentos permitem ao sistema estimar uma probabilidade de fraude para cada transação através de uma rede neural. A abordagem também produz explicações para uma determinada previsão, uma vez que também pode emitir certos códigos expressando motivos que revelam a contribuição relativa de cada característica para um determinado resultado, e ser robusta à desvio de conceito, visto que seu desempenho é monitorado e o modelo é re-treinado quando cai abaixo de um limite predeterminado.

\noindent
\textbf{Detecção de Fraude por Aprendizado baseado em Grafos.} \cite{weber2018scalable} é um dos poucos trabalhos atuais que aborda a detecção de fraude usando técnicas de aprendizagem de grafos, como a construção de Redes Convolucionais de Grafos, um tipo recente de redes neurais profundas, introduzidas por \cite{kipf2016semi, atwood2016diffusion} e posteriormente aprimoradas em sua escalabilidade por \cite{chen2018fastgcn}. Em \cite{weber2018scalable}, os autores propõem duas arquiteturas distintas para lidar com o problema de lavagem de dinheiro usando uma GCN (Graph Convolutional Network) e uma FastGCN (Fast Graph Convolutional Networks) para serem treinadas em um conjunto de dados sintético (AMLSim). Esse conjunto de dados simula uma série de transações bancárias juntamente com um conjunto de padrões de lavagem de dinheiro suspeitos, fazendo com que algumas features geradas sinteticamente se comportem como outliers, como o valor total da transação sendo muito maior do que a média. Eles também concluíram que a arquitetura FastGCN é mais adequada para este conjunto de dados específico pelo fato de conseguir ser treinada em menos tempo para uma quantidade fixa de épocas.

Depois desse estudo inicial, os autores dessa pesquisa decidiram abordar os cenários de lavagem de dinheiro em Bitcoin usando técnicas de aprendizagem baseadas em grafos sofisticadas \cite{weber2019anti}, como Skip-GCN, uma Rede Convolucional de Grafos contendo conexões de salto via compartilhamento de pesos de aresta entre níveis de profundidade distintos da rede neural, e EvolveGCN, por \cite{pareja2020evolvegcn}, uma arquitetura de rede convolucional baseados em grafos que lida com as mudanças temporais comuns desses tipos de conjuntos de dados (desvio de conceito) usando técnicas de memória de longo prazo. Eles concluíram em \cite{weber2019anti} que a detecção de fraude é um problema difícil por suas características complexas intrínsecas, dado que um modelo Random Forest tradicional foi capaz de superar todos os métodos propostos de aprendizagem baseados em grafos apenas dando-lhe as mesmas características, levantando a possibilidade de um futuro aprimoramento do modelo. A partir disso, foi criado um conjunto de GCNs para imitar o comportamento de uma floresta aleatória. Porém, os autores também mostraram que esses novos modelos (Skip-GCN e EvolveGCN) são capazes de atingir um desempenho melhor do que sua antecessora, a GCN.

Um estudo mais recente \cite{liu2021pickandchoose} realizou um avanço significativo ao conseguir adaptar e utilizar redes neurais baseadas em grafos para conjuntos de dados altamente desbalanceados aplicados ao contexto de detecção de fraudes em dados em rede. Nesse estudo é realizado uma etapa de amostragem nos vértices do grafo de forma a equilibrar o número de vértices pertencentes a cada classe e facilitar a propagação do sinal da classe minoritária aos vértices mais próximos. Utilizando essa estratégia, os autores concluíram que essa estratégia melhora efetivamente o desempenhos dos modelos quando aplicados a esse contexto de problema.

\noindent
\textbf{Relação deste trabalho com o estado da arte.} Este artigo compara o desempenho de abordagens tradicionais com o de abordagens baseadas em grafos na detecção de fraudes financeiras, levando em consideração o desbalanceamento de classes e o desvio de conceito, que são características inerentes deste problema.

\section{Metodologia}
\label{methodology}

Nesta seção apresentamos os conjuntos de dados, modelos e métricas aqui utilizados.

\subsection{Conjuntos de Dados}

Para esse estudo, utilizamos alguns conjuntos de dados disponíveis publicamente.

\begin{table}[t]
\centering
\footnotesize
\begin{tabular}{c|cccccc}
\toprule
\begin{tabular}[c]{@{}c@{}}Conjunto de\\ Dados\end{tabular} & \begin{tabular}[c]{@{}c@{}}Transações\\ Legítimas\end{tabular} & \begin{tabular}[c]{@{}c@{}}Transações\\ Ilícitas\end{tabular} & \begin{tabular}[c]{@{}c@{}}Usuários\\ Legítimos\end{tabular} & \begin{tabular}[c]{@{}c@{}}Usuários\\ Ilícitos\end{tabular} & TD & \begin{tabular}[c]{@{}c@{}}Desvio de\\ Conceito?\end{tabular} \\ \midrule
AMLSim 1 & 5000 & 5000 & 500 & 500 & 1 & Não \\
AMLSim 2 & 9500 & 500 & 950 & 50 & 19 & Não \\
AMLSim 3 & 9500 & 500 & 950 & 50 & 19 & Sim \\ \bottomrule
\end{tabular}
\caption{Estatísticas do conjunto de dados sintético AMLSim.}
\label{amlsim_data_table}
\end{table}

\noindent
\textbf{Dados sintéticos gerados com \textit{Anti-money Laundering Simulator} (AMLSim):} O AMLSim\footnote{\url{https://github.com/IBM/AMLSim/}} \cite{pareja2020evolvegcn, weber2018scalable} é um simulador multiagente proposto para gerar conjunto de dados para o domínio de detecção de fraudes baseado em outro simulador de transações mais genérico (não possui nenhuma especificidade para gerar comportamentos de fraude), denominado PaySim \cite{lopez2016paysim}. A partir do AMLSim foram gerados três conjuntos de dados, conforme demonstrado na Tabela~\ref{amlsim_data_table}, com o objetivo de estudar a robustez dos métodos existentes em relação ao desbalanceamento de classe e ao desvio de conceito. Esses conjuntos de dados são gerados baseados em diferentes arquivos de parametrização, os quais são utilizados para gerar dados que imitam diversas transações legítimas e fraudulentas.

\begin{figure}[t]
    \centering
    \includegraphics[width=0.5\textwidth]{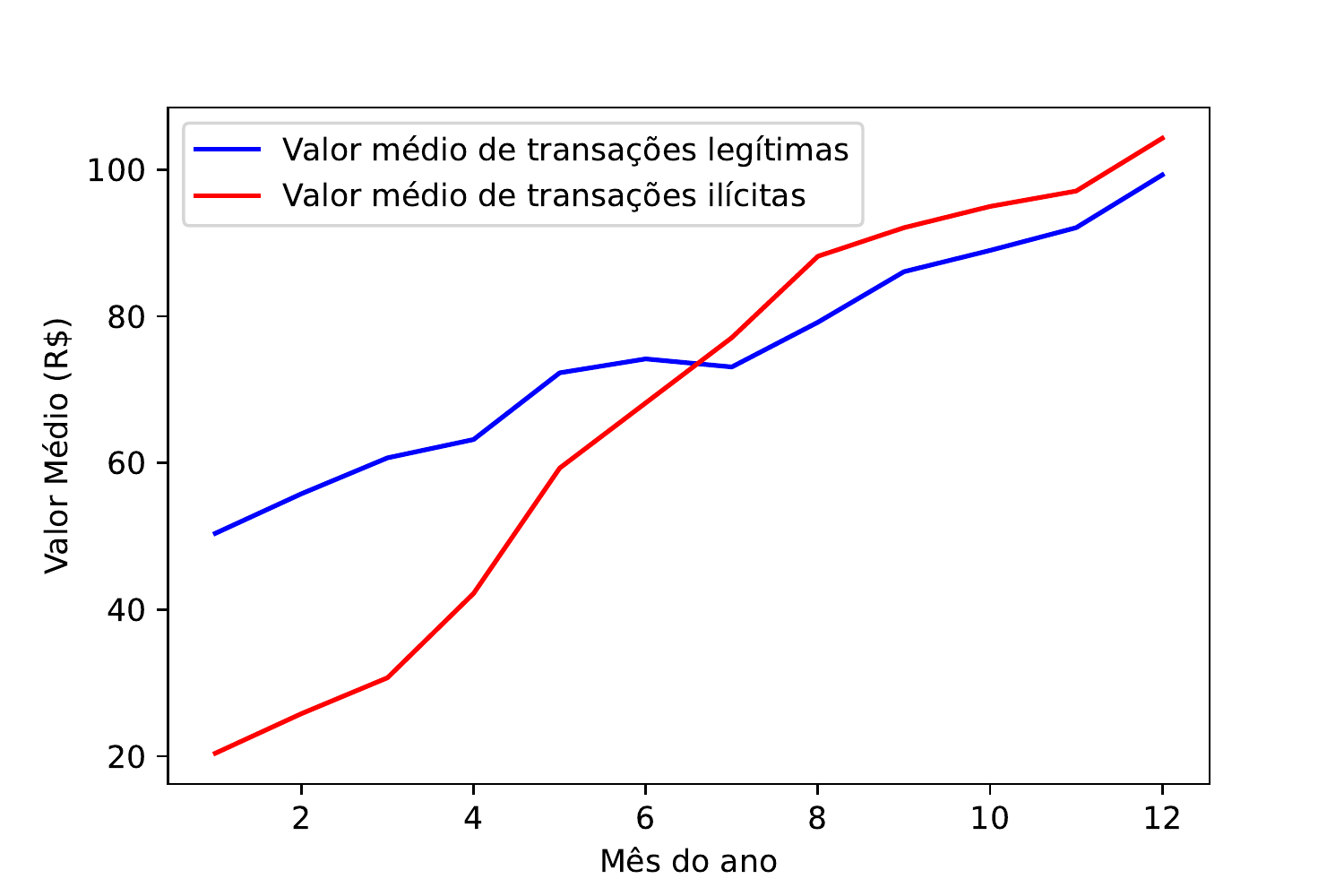}
    \caption{Variação do valor médio em dinheiro de transações legítimas e ilícitas ao longo de um ano para o AMLSim 3.}
    \label{fig:concept_drift}
\end{figure}

O primeiro conjunto de dados, AMLSim 1, é perfeitamente balanceado e servirá como um \textit{baseline} para os demais. O segundo, AMLSim 2, possui um forte desbalanceamento de classes e servirá para demonstrar o impacto desse fenômeno no desempenho dos modelos estudados. O terceiro, AMLSim 3, também foi gerado com uma alta taxa de desbalanceamento entre as classes, porém com um efeito adicional de desvio de conceito, definido na Seção \ref{concept_drift}. Mais precisamente, os valores médios das transações legítimas e ilícitas é variado mês-a-mês, durante 12 meses de simulação, conforme pode-se observar na Figura \ref{fig:concept_drift}. Observe ainda que os valores médios chegam a se cruzar após o mês 7.

\noindent    
\textbf{Elliptic Data Set:} é um conjunto de dados coletado durante cerca de duas semanas diretamente do \textit{blockchain} da criptomoeda Bitcoin e mapeia transações de entidades reais anonimizadas como legítimos (bolsas de valores, fornecedores de carteira, mineradores, serviços lícitos, etc) ou como ilícitos (golpes, \textit{malware}, organizações terroristas, \textit{ramsomware}, etc)  \cite{weber2019anti}. Ele contém 203769 vértices (transações) e 234355 arestas (operações financeiras). Dentre os vértices, apenas 23\% desses vértices são rotulados, sendo 2\% ilícitos (4545 nós) e 21\% (42019 nós) legítimos, enquanto os 77\% restantes não possuem rotulação alguma.

Existem 166 características associadas a cada nó do grafo, porém, devido a questões de privacidade, não foi fornecida uma descrição exata de todas elas. O tempo de observação foi discretizado em 49 janelas uniformemente espaçadas. Cada nó está associado a uma única janela de tempo e as suas arestas representam cada operação de transferência, depósito ou retirada, representando uma medida no tempo em que uma transação foi transmitida para a rede de Bitcoin. Cada janela de tempo contém um único componente conexo de transações que apareceram no \textit{blockchain} em menos de três horas entre si e não há arestas conectando as diferentes etapas de tempo.

As primeiras 94 características representam informações locais sobre a transação, incluindo a etapa de tempo descrita acima, número de entradas e saídas, taxa de transação, volume de saída e outros números agregados, como média de Bitcoins recebidos e gastos. As 72 características restantes são agregações obtidas usando informações de transação de saltos entre o nó central, dando o máximo, mínimo, desvio padrão e coeficientes de correlação das transações vizinhas para os mesmos dados de informação (número de entradas/saídas, taxa de transação, etc).

\noindent
\textbf{YelpCHI:} é um conjunto de dados coletado do site yelp.com que consiste em 67395 avaliações para um conjunto de 201 hotéis e restaurantes na área de Chicago e contém informações dos produtos e de seus 38063 usuários, carimbos de data e hora, classificações e a avaliação em texto livre \cite{mukherjee2013yelp, rayana2015collective, rayana2016collective}.
O Yelp possui um algoritmo interno de filtragem  que identifica avaliações suspeitas e as guarda em uma lista separada das avaliações legítimas. Embora esse filtro não seja perfeito, ele produziu resultados precisos em estudos realizados anteriormente \cite{weise2011lie}. Portanto, seus resultados podem ser usados como rótulos para determinadas avaliações.

Nesse conjunto de dados, possuímos 58479 (87,77\%) avaliações legítimas, 8916 (13,23\%) avaliações filtradas, 30325 (79,67\%) usuários legítimos e 7738 (20,33\%) usuários suspeitos. Desse modo, a taxa de desbalanceamento dos rótulos de usuários é de 3,92, indicando um desbalanceamento moderado, quando comparado com os demais datasets utilizados nessa pesquisa. Para a montagem do grafo, foram utilizadas somente as relações de Usuário-Avaliação-Produto (U-A-P), de modo que cada usuário seja ligado às suas avaliações de cada produto por meio de arestas.
 
\subsection{Modelos}
\label{models}

Nos experimentos, utilizamos algumas implementações de modelos clássicos, que não são capazes de incorporar informação sobre a relação entre os nós:

\begin{itemize}
    \item XGBoost \cite{chen2016xgboost};
    \item CatBoost \cite{dorogush2018catboost}.
\end{itemize}
além de alguns modelos de redes neurais baseadas em grafos como baselines:

\begin{itemize}
    \item GCN \cite{kipf2016semi}: utiliza aproximação local de primeira ordem da convolução de grafos espectrais;
    \item GAT \cite{velivckovic2017graph}: utiliza mecanismo de atenção para agregação de vizinhos;
    \item GraphSAGE \cite{hamilton2017inductive}: utiliza indução em uma amostra de tamanho fixo dos nós vizinhos;
    \item GraphSAINT \cite{zeng2019graphsaint}: utilizando amostragem dos nós do grafo, para maior escalabilidade e eficiência.
\end{itemize}
e, por último, alguns modelos de redes neurais baseadas em grafos como estado da arte:

\begin{itemize}
    \item EvolveGCN \cite{pareja2020evolvegcn}: usa uma combinação de convoluções em grafos com técnicas de memória de longo prazo para lidar com as mudanças temporais gradativas nos dados (desvio de conceito);
    \item PC-GNN \cite{liu2021pickandchoose}: utiliza de um novo método de amostragem para realizar a propagação do sinal da classe minoritária aos nós vizinhos. Isso é feito em três etapas: amostragem, escolha e agregação. Os nós centrais são amostrados com um balanceador de classes de modo a construir um sub-grafo balanceado para a etapa de treinamento. Sob uma medida de distância parametrizada, a vizinhança de cada nó pertencente à classe minoritária é sobreamostrada, enquanto a vizinhança de cada nó pertencente à classe majoritária é subamostrada. Em seguida, faz-se uma agregação das propagações de mensagens de diferentes sub-grafos para se obter uma representação final do grafo completo.
\end{itemize}

Todas essas implementações estão disponíveis e foram compiladas por mim em um repositório público\footnote{\url{https://github.com/ronaldpereira/brasnam-experiments}}.

\subsection{Métricas}

Para medir e comparar o desempenho dos modelos, utilizamos duas métricas amplamente adotadas em problemas de classificação, F1-macro e AUC, além de uma métrica proposta especificamente para problemas de detecção de fraude, denominada F1-fraud.

A F1-macro é a média aritmética simples do F1-score de cada classe. Já a F1-fraud é o escore F1 considerando como classe positiva somente as instâncias rotuladas como fraudes. Embora correlacionada com a primeira, esta métrica dá ênfase às dimensões precisão e revocação especificamente para classe de maior interesse. A terceira métrica é a área sob a curva ROC (AUC) e indica a probabilidade que uma instância positiva aleatória do conjunto de teste seja rotulada como mais provável de ser fraudulenta que uma instância negativa aleatória do mesmo conjunto. Uma característica importante do AUC é que ele é independente do limiar de probabilidade escolhido para classificar instâncias como negativas ou positivas.

\section{Resultados}
\label{results}

Nesta seção iremos descrever os resultados dos experimentos realizados nessa pesquisa.

\subsection{Avaliação Geral dos Métodos}

Para mensurar corretamente a efetividade de cada método na classificação de nós em suas respectivas classes, foram utilizados em todos os conjuntos de dados (exceto AMLSim 3) uma separação estratificada, de modo a manter a taxa de desbalanceamento constante entre esses conjuntos, em 60\% treino, 20\% validação e 20\% teste.

A Tabela~\ref{tab:results_yelp_elliptic} mostra o desempenho de cada método nos dois conjuntos de dados reais (YelpChi e Elliptic Data Set) em função das três métricas de avaliação. Os métodos mais recentes, tidos como estado da arte - PC-GNN e EvolveGCN -, se mostraram mais efetivos do que os demais. Com esse resultado, também foi possível concluir também que a nossa implementação do PC-GNN também está bem próxima dos resultados apresentados pelo artigo original, atestando a corretude da nossa implementação.

\begin{table}[t]
\resizebox{\textwidth}{!}{
\centering
\begin{tabular}{c|ccc|ccc}
\toprule
\begin{tabular}[c]{@{}c@{}}Conjunto\\ de Dados\end{tabular} & \multicolumn{3}{c|}{YelpChi} & \multicolumn{3}{c}{Elliptic Data Set} \\ \midrule
Métrica & F1-macro & F1-fraud & AUC & F1-macro & F1-fraud & AUC \\ \midrule
XGBoost & ,546 $\pm$ ,054 & ,261 $\pm$ ,027 & ,655 $\pm$ ,045 & ,564 $\pm$ ,058 & ,259 $\pm$ ,026 & ,583 $\pm$ ,043 \\  
CatBoost & ,518 $\pm$ ,055 & ,213 $\pm$ ,026 & ,654 $\pm$ ,046 & ,543 $\pm$ ,059 & ,201 $\pm$ ,025 & ,571 $\pm$ ,042 \\  \midrule
GCN & ,576 $\pm$ ,059 & ,245 $\pm$ ,028 & ,620 $\pm$ ,042 & ,458 $\pm$ ,041 & ,044 $\pm$ ,024 & ,590 $\pm$ ,037 \\  
GAT & ,551 $\pm$ ,059 & ,117 $\pm$ ,025 & ,643 $\pm$ ,045 & ,463 $\pm$ ,045 & ,026 $\pm$ ,024 & ,527 $\pm$ ,035 \\  
GraphSAGE & ,471 $\pm$ ,054 & ,108 $\pm$ ,025 & ,579 $\pm$ ,040 & ,434 $\pm$ ,045 & ,055 $\pm$ ,022 & ,542 $\pm$ ,036 \\  
GraphSAINT & ,605 $\pm$ ,067 & ,282 $\pm$ ,034 & ,727 $\pm$ ,052 & ,568 $\pm$ ,060 & \textbf{,378 $\pm$ ,027} & ,669 $\pm$ ,044 \\ \midrule
EvolveGCN & \textbf{,667 $\pm$ ,067} & \textbf{,459 $\pm$ ,033} & \textbf{,756 $\pm$ ,051} & \textbf{,623 $\pm$ ,064} & \textbf{,285 $\pm$ ,031} & \textbf{,744 $\pm$ ,049} \\ 
PC-GNN & \textbf{,664 $\pm$ ,063} & \textbf{,370 $\pm$ ,032} & \textbf{,757 $\pm$ ,049} & \textbf{,612 $\pm$ ,060} & ,259 $\pm$ ,032 & \textbf{,699 $\pm$ ,049} \\ \bottomrule
\end{tabular}
}
\caption{Desempenho dos modelos nos conjuntos de dados reais.}
\label{tab:results_yelp_elliptic}
\end{table}

\subsection{Degradação em função do Desbalanceamento de Classes}

Para avaliar a robustez dos métodos em relação ao desbalanceamento de classes, realizamos experimentos utilizando os dois conjuntos de dados sintéticos, AMLSim 1 e AMLSim 2, cujas taxas de desbalanceamento TD são iguais a 1 (perfeitamente balanceado) e 19 (extremamente desbalanceado), respectivamente.

A Tabela~\ref{tab:results_amlsim_1_2} apresenta o desempenho de cada método para estes datasets. Analisando apenas o AMLSim 1, observa-se que os modelos do estado da arte para detecção de fraude podem ser superados pelos outros métodos quando os dados estão balanceados. Por outro lado, os resultados no AMLSim 2 mostram que quando há desbalanceamento, todos os métodos, com exceção do EvolveGCN e do PC-GNN, apresentam um baixíssimo desempenho, com AUC muito próximo de 0,5. Surpreendentemente, o EvolveGCN obteve desempenho significativamente superior (do ponto de vista estatístico) neste cenário do que no cenário balanceado, em relação às métricas F1-macro e F1-fraud.

\begin{table}[t]
\resizebox{\textwidth}{!}{
\centering
\begin{tabular}{c|ccc|ccc}
\toprule
\begin{tabular}[c]{@{}c@{}}Conjunto\\ de Dados\end{tabular} & \multicolumn{3}{c|}{AMLSim 1 (TD = 1)} & \multicolumn{3}{c}{AMLSim 2 (TD = 19)} \\ \midrule
Métrica & F1-macro & F1-fraud & AUC & F1-macro & F1-fraud & AUC \\ \midrule
XGBoost & ,841 $\pm$ ,091 & ,906 $\pm$ ,041 & ,939 $\pm$ ,067 & ,440 $\pm$ ,045 & ,054 $\pm$ ,021 & ,535 $\pm$ ,037 \\ 
CatBoost & \textbf{,865 $\pm$ ,082} & \textbf{,956 $\pm$ ,042} & \textbf{,964 $\pm$ ,057} & ,473 $\pm$ ,047 & ,070 $\pm$ ,022 & ,567 $\pm$ ,035 \\ \midrule
GCN & ,840 $\pm$ ,083 & ,714 $\pm$ ,041 & ,910 $\pm$ ,062 & ,497 $\pm$ ,048 & ,242 $\pm$ ,023 & ,554 $\pm$ ,035 \\ 
GAT & ,833 $\pm$ ,080 & ,712 $\pm$ ,041 & ,932 $\pm$ ,061 & ,461 $\pm$ ,044 & ,008 $\pm$ ,020 & ,542 $\pm$ ,039 \\ 
GraphSAGE & \textbf{,874 $\pm$ ,089} & ,825 $\pm$ ,043 & \textbf{,994 $\pm$ ,066} & ,512 $\pm$ ,052 & ,127 $\pm$ ,029 & ,620 $\pm$ ,044 \\ 
GraphSAINT & ,840 $\pm$ ,082 & \textbf{,975 $\pm$ ,041} & ,947 $\pm$ ,062 & ,566 $\pm$ ,052 & ,292 $\pm$ ,029 & ,616 $\pm$ ,045 \\ \midrule
EvolveGCN & ,718 $\pm$ ,074 & ,642 $\pm$ ,037 & ,864 $\pm$ ,054 & \textbf{,781 $\pm$ ,072} & \textbf{,826 $\pm$ ,037} & \textbf{,863 $\pm$ ,054} \\ 
PC-GNN & ,748 $\pm$ ,076 & ,529 $\pm$ ,037 & ,854 $\pm$ ,056 & \textbf{,747 $\pm$ ,077} & \textbf{,762 $\pm$ ,038} & \textbf{,831 $\pm$ ,057} \\ \bottomrule
\end{tabular}
}
\caption{Desempenho dos modelos no experimento sobre desbalanceamento.}
\label{tab:results_amlsim_1_2}
\end{table}
 
\subsection{Degradação ao longo do tempo devido ao Desvio de Conceito}

Para avaliar a robustez dos métodos a um desvio de conceito gradual de classes, realizamos experimentos utilizando o conjunto de dados sintético AMLSim 3. Utilizamos os meses 1 a 4 para definir o conjunto de dados de treinamento, enquanto os dados restantes foram usados para definir dois conjuntos de teste: o primeiro contendo os meses 5 a 8, e o segundo contendo os meses 9 a 12. Conforme explicado na Seção \ref{concept_drift}, esses dados sofrem de uma mudança na distribuição de suas características mês a mês, de modo que quanto maior a distância entre os meses de treinamento e os meses de teste, menor o desempenho esperado do modelo.

A Tabela~\ref{tab:results_amlsim_2_3} mostra os resultados obtidos para os dois conjuntos de teste. Analisando os resultados para o primeiro conjunto (meses 5 a 9), observa-se que os métodos tiveram um baixo desempenho geral em termos de F1-macro e F1-fraud, mas o AUC obtido por alguns modelos se manteve próximo ou superior a 0,7. A explicação para isto é que, embora o limiar de classificação definido com base no conjunto de treinamento cometa muitos erros nos casos de teste, o ranqueamento dos nós segundo a probabilidade de serem fraudulentos continua razoavelmente boa, fazendo com que o AUC não se degrade muito. Neste caso, um sistema de detecção de fraude poderia ser ``corrigido'' com um simples ajuste deste limiar.

Por outro lado, os resultados para a segundo conjunto (meses 10 a 12) mostram um resultado bem mais negativo. Além da queda ainda mais acentuada no F1-macro e F1-fraud, o AUC de todos os métodos, com exceção do EvolveGCN e do PC-GNN ficam abaixo de 0,4, indicando que tais métodos costumam classificar instâncias negativas como pais prováveis de serem fraudes do que as instâncias positivas. Isto é explicado pelo cruzamento das curvas na Figura \ref{fig:concept_drift}. Desse modo, nenhum dos métodos do estado da arte foram capazes de manter um desempenho razoável quando expostos ao desvio de conceito, explicitando a necessidade de se realizar progressos nessa vertente.

\begin{table}
\resizebox{\textwidth}{!}{
\centering
\begin{tabular}{c|ccc|ccc}
\toprule
\begin{tabular}[c]{@{}c@{}}Conjunto\\ de Dados\end{tabular} & \multicolumn{3}{c|}{\begin{tabular}[c]{@{}c@{}}AMLSim 3 (TD = 19)\\ (Meses 5, 6, 7 e 8)\end{tabular}} & \multicolumn{3}{c}{\begin{tabular}[c]{@{}c@{}}AMLSim 3 (TD = 19)\\ (Meses 9, 10, 11 e 12)\end{tabular}} \\ \midrule
Métrica & F1-macro & F1-fraud & AUC & F1-macro & F1-fraud & AUC \\ \midrule
XGBoost & ,419 $\pm$ ,056 & ,390 $\pm$ ,027 & ,698 $\pm$ ,045 & ,287 $\pm$ ,026 & ,158 $\pm$ ,012 & ,321 $\pm$ ,025 \\ 
CatBoost & ,434 $\pm$ ,052 & ,398  $\pm$ ,032 & ,699 $\pm$ ,057 & ,365 $\pm$ ,029 & ,048 $\pm$ ,011 & ,356 $\pm$ ,026 \\\midrule
GCN & ,432 $\pm$ ,063 & ,188 $\pm$ ,032 & ,542 $\pm$ ,047 & ,269 $\pm$ ,048 & ,000 $\pm$ ,000 & ,339 $\pm$ ,039 \\ 
GAT & ,379 $\pm$ ,052 & ,107 $\pm$ ,027 & ,508 $\pm$ ,043 & ,166 $\pm$ ,052 & ,000 $\pm$ ,000 & ,298 $\pm$ ,042 \\ 
GraphSAGE & ,491 $\pm$ ,065 & ,305 $\pm$ ,031 & ,697 $\pm$ ,048 & ,328 $\pm$ ,055 & ,032 $\pm$ ,023 & ,435 $\pm$ ,041 \\ 
GraphSAINT & ,534 $\pm$ ,062 & ,277 $\pm$ ,031 & ,707 $\pm$ ,049 & ,317 $\pm$ ,040 & \textbf{,156 $\pm$ ,020} & ,498 $\pm$ ,033 \\ \midrule
EvolveGCN & \textbf{,568 $\pm$ ,069} & \textbf{,425 $\pm$ ,032} & \textbf{,711 $\pm$ ,050} & \textbf{,453 $\pm$ ,048} & \textbf{,292 $\pm$ ,024} & \textbf{,579 $\pm$ ,035} \\ 
PC-GNN & \textbf{,684 $\pm$ ,065} & \textbf{,532 $\pm$ ,033} & \textbf{,777 $\pm$ ,052} & \textbf{,469 $\pm$ ,054} & ,129 $\pm$ ,026 & \textbf{,547 $\pm$ ,042} \\ \bottomrule
\end{tabular}
}
\caption{Desempenho dos modelos no experimento sobre desvio de conceito.}
\label{tab:results_amlsim_2_3}
\end{table}

\section{Conclusões e Trabalhos Futuros}
\label{conclusion}

Nesse trabalho, propusemos uma nova metodologia para se avaliar a robustez de métodos de detecção de fraudes financeiras baseadas em geração de dados sintéticos utilizando o AMLSim. Desse modo, conseguimos visualizar os pontos fortes e pontos fracos de cada um dos modelos anteriormente apresentados. Também utilizando conjuntos de dados reais, mostramos que técnicas propostas recentemente, como o EvolveGCN e o PC-GNN alcançam um excelente desempenho na tarefa de classificação, a despeito da complexidade inerente ao problema. No entanto, a combinação das duas características que distinguem a detecção de fraudes de tarefas comuns de classificação -- o desbalanceamento de classe e o desvio de conceito -- ainda impõe desafios para estas técnicas. Concluímos que ainda não existe um modelo que seja capaz de lidar com estas duas características simultaneamente. Como direções futuras, acreditamos que técnicas que utilizem amostragem para compensar o desbalanceamento e que incorporem, ao mesmo tempo, um modelo para a dinâmica da relação entre \textit{features} e o rótulo possam ser promissoras para o avanço na área de detecção de fraudes para dados em rede.

% \bibliographystyle{sbc}
% \bibliography{body/references}

\end{document}